\documentclass{article}
\usepackage[utf8]{inputenc} 
\usepackage[T1]{fontenc} 
\usepackage{hyperref} 
\usepackage{url} 
\usepackage{booktabs} 
\usepackage{amsfonts} 
\usepackage{nicefrac} 
\usepackage{microtype} 
\usepackage{xcolor}
\usepackage{times}
\usepackage{natbib}
\usepackage{amsmath}
\usepackage{color}
\usepackage{authblk}
\usepackage{enumitem}

\title{AI Scientists as Engines of Discovery: A Case for Development within Reformed Institutions}

\author[1,2]{Raul Jimenez}
\author[3,4]{Boris Bolliet}
\author[5,6]{Francisco Villaescusa-Navarro}
\author[7]{Rabih Zbib}
\author[8,9]{Benjamin Wandelt}
\author[10]{David N. Spergel}
\author[11]{Thomas Meier}
\author[12]{Jessica Montgomery}
\author[12,13]{Hana Aliee}
\author[1,2]{Licia Verde}

\affil[1]{Institute of Cosmos Sciences (ICCUB), University of Barcelona, Barcelona, Spain}
\affil[2]{ICREA, Barcelona, 08010, Spain}
\affil[3]{Cavendish Astrophysics, University of Cambridge, Madingley Road, Cambridge CB3 0HA, UK}
\affil[4]{Kavli Institute for Cosmology, University of Cambridge, Madingley Road, Cambridge CB3 0HA, UK}
\affil[5]{Center for Computational Astrophysics, Flatiron Institute, New York, NY 10010, USA}
\affil[6]{Department of Astrophysical Sciences, Princeton University, Princeton, NJ 08544, USA}
\affil[7]{Avature, 535 5th Avenue, Suite 1500, New York, NY 10017, USA}
\affil[8]{Department of Physics and Astronomy, Johns Hopkins University, Baltimore, MD 21218, USA}
\affil[9]{Department of Applied Mathematics and Statistics, Johns Hopkins University, Baltimore, MD 21218, USA}
\affil[10]{Flatiron Institute, New York, NY 10010, USA}
\affil[11]{Munich Center for Machine Learning, LMU Munich, Munich, Germany}
\affil[12]{Department of Computer Science and Technology, University of Cambridge, Cambridge, CB3 0FD, UK}
\affil[13]{School of Clinical Medicine, University of Cambridge, Cambridge, CB3 0FD, UK}

\date{\today}

\begin{document}

\maketitle

\begin{abstract}
Agentic artificial intelligence (AI) systems are beginning to assist, accelerate, and partially automate scientific discovery, performing tasks that span literature synthesis, code generation, data analysis, hypothesis proposal, and model criticism. We argue that this transition is qualitative rather than incremental, and that suitably designed multi-agent systems may evolve from passive computational tools into ``AI scientists'' that can expand the hypothesis-generating and verification capacity of science. 
Such systems must be developed and deployed within a scientific ecosystem fit for purpose: institutions must be redesigned for verification, accountability, interpretability, and dual-use safety. We sketch how multi-agent architectures, illustrated by the prototype framework \textit{Denario}, 
accelerate the discovery cycle and traverse model spaces beyond human reach; examine what this implies for authorship, peer review, and the enduring role of human scientists; and close with recommendations for governing AI as an epistemic actor rather than a mere instrument.
\end{abstract}

Scientific discovery has always evolved alongside technological innovation. New instruments expand the reach of human perception; new mathematical and computational tools expand reasoning. The telescope allowed Galileo to discover the moons of Jupiter in 1610; the microscope opened the biological world; the transistor laid the foundation for the digital age; differential geometry gave Einstein the language of general relativity. Each development altered the scale and speed of progress, yet none replaced the scientist: they amplified human capability and expanded human ingenuity.

Artificial intelligence is positioned as the next such transformation, but with a difference. Earlier instruments extended perception or accelerated computation; current AI systems are beginning to participate in the cognitive operations of science itself, proposing models, criticizing them, and revising assumptions in response to discrepancies. The emergence of powerful machine learning systems has sparked excitement and anxiety ~\citep{trotta2025ai,Kusumegi2026,Zhao:2026hallucinations,peiris2026llm,Bertone2026}, with some observers fearing that algorithms capable of writing code, analyzing data, and generating hypotheses may diminish the role of human researchers or make scientific abilities atrophy.

Science has long been shaped not only by the technologies it uses but also by the norms, structures, and institutions built around them. The printing press created new forms of scientific communication; digital computing changed what could be calculated and how results were shared. In each case, technological capability brought new governance needs, and the governance response was part of the technology's development. The red flag walking ahead of a locomotive has become technologist shorthand for the idea that regulation delays adoption. In many domains, however, the right governance with the right incentives makes innovation usable at scale. This includes traffic codes and driver licensing for motoring, spectrum allocation for mobile telephony, pharmaceutical regulation, and financial legislation. This invites a question about how to equip the institutions of science to steward AI for scientific and social benefit.

Scientific discovery is increasingly limited by a cognitive bottleneck of information processing, hypothesis generation, validation and synthesis. Current AI systems, used in isolation, do not yet break this bottleneck: they assist but do not \emph{reason}, they generate but do not \emph{falsify} reliably, they interpolate but do not \emph{extrapolate}. However, a different framework is now emerging in which AI multi-agent systems become \emph{epistemic actors} embedded within the scientific method. We will call such systems \emph{AI scientists}. We are not yet there, but the trajectory is qualitative rather than incremental.

The central thesis is that, because they can expand the hypothesis-generating and verification capacity of science, such systems must be developed and deployed within institutions redesigned around verification, accountability, interpretability, and dual-use safety. Science is a collective attempt to discover how the world works, and using tools to make that discovery faster can accelerate human capabilities. The argument proceeds in five steps: (i)~what current AI systems actually do; (ii)~how multi-agent architectures function as algorithmic research groups, illustrated by the prototype \textit{Denario}; (iii)~how they 
accelerate the discovery cycle and expand the explorable hypothesis space; (iv)~what enduring role this leaves for human scientists; and (v)~what institutional, structural and systemic changes are required.

\section*{What current AI systems actually do}

Large language models (LLMs) form the foundation of the current wave of artificial intelligence. They are statistical systems that predict sequences of tokens---words, symbols, or text fragments---from patterns learned in extremely large datasets. Modern models rely on transformer architectures, embedding tokens in high-dimensional vector spaces and using self-attention to evaluate how strongly each token influences the interpretation of others. The model thereby learns to generate coherent continuations of text, code, or mathematical expressions.

These systems do not ``understand'' concepts in the way humans do. They encode statistical relationships across vast bodies of information; and yet, to encode such relationships efficiently they reconstruct in their embeddings concepts we recognize as natively human. This makes LLMs powerful tools for synthesis: they can summarize research papers, generate computer programs, suggest mathematical derivations, and organize datasets. What they currently fail to do reliably, in isolation, is reason in a sustained way. However, AI {\it agents}, by using tools, planning and performing series of actions, maintaining state over sequences of actions, and adapting toward a goal, can overcome this limitation. The most transformative scientific applications will therefore not arise from prompting a model alone, but from systems composed of multiple cooperating agents.

\section*{Multi-agent AI as an algorithmic research group: the Denario case}

Shortly after LLMs 
became prominent, researchers recognized that the most powerful applications would involve coordinated systems of specialized agents rather than isolated calls to a single model. Fine-tuning was rapidly complemented by context engineering and scaffolding with external tools. In such architectures some components retrieve literature and contextual knowledge; others generate and execute code; others analyze data, visualize results, or evaluate candidate models. A reasoning layer coordinates the components by deciding which tools to invoke and how to interpret their outputs.

For instance, a useful conceptual decomposition may assign four roles: a \emph{generation agent} proposes models or interpretations; a \emph{critic agent} identifies inconsistencies, overfitting, or violations of constraints; a \emph{verification agent} tests predictions against data; and a \emph{controller agent} coordinates the others. This architecture enables an internal dialectic in which competing ideas are explored and refined, and can generate novel insights by traversing regions of model parameter space that human researchers may never have considered~\citep{ghafarollahi2024sciagents,lu2024aiscientist,yamada2025aiscientistv2,gottweis2025aicoscientist}. The system resembles a research team, with the crucial difference that the collaborators are algorithmic agents operating at scale and at machine speed.

To probe this paradigm concretely, some of us have developed \textit{Denario}~\citep{villaescusa2025denario}, a multi-agent framework for scientific discovery.\footnote{\url{https://astropilot-ai.github.io/DenarioPaperPage/}} Denario integrates automated literature exploration, code generation and execution, dataset analysis, and iterative hypothesis testing; cooperating agents communicate through structured prompts and share computational tools, while a central reasoning layer orchestrates the workflow. In one controlled demonstration~\citep{braga2026dhost} Denario was asked to unveil the hidden symmetries that keep certain modified-gravity theories ``healthy'' (ghost-free), and it identified them on its own. The exercise illustrates a broader possibility: AI systems can search candidate model spaces far faster than humans can, allowing scientists to focus on interpretation and conceptual reasoning.

\section*{Compressing the discovery cycle and expanding the hypothesis space}

Scientific progress typically follows a cycle: formulating hypotheses, developing computational or experimental tools, analyzing data, and communicating results~\citep{popper1963conjectures}. Historically, each stage required substantial effort. Multi-agent AI systems compress several stages at once: literature retrieval is accelerated, analysis code can be generated and refactored almost instantaneously, and numerical experiments can be orchestrated dynamically. Even where physical experimentation remains the rate-limiting step, AI can examine prior experimental results and reconcile them with proposed hypotheses at a scale no individual investigator could match, sharpening the design of the next experiment.

The deeper change concerns the hypothesis space itself. Many scientific problems involve enormous spaces of possible models, for example, families of gravitational theories or inflationary scenarios in cosmology, and combinatorial possibilities of genetic interactions in biology. Human intuition guides exploration toward promising subsets; this has produced remarkable successes, but leaves most of the landscape unexplored. AI systems can generate candidate models, evaluate them against data, identify promising contenders, and propose new experiments to discriminate among them at scale. In effect they are engines for navigating scientific landscapes at super-human scale, changing not only the speed of science but also, we argue here, its reach.

\section*{The enduring role of human scientists}

Despite this trajectory, human scientists will remain essential. Scientific revolutions rarely arise from incremental advances alone. Newton unified celestial and terrestrial mechanics; Maxwell unified electricity and magnetism; Einstein reshaped our understanding of space and time. These breakthroughs involved imaginative leaps rather than simple extensions of existing data. Paradigm shifts of this magnitude are rare; most progress is steady and incremental, consisting in the methodical retrieval and synthesis of literature, experiments and observations. AI excels at the latter, and at spanning areas of expertise no single human can command. But the framing of meaningful questions, and the recognition of when a result matters and what it means, remains deeply human.

It is useful to think in terms of Kauffman's "adjacent possible"~\citep{kauffman2000investigations}. 
AI is superb at exploring the adjacent possible: mapping, extending, combining, and working through possibilities that are already latent in the existing structure of knowledge. It can do this at unprecedented scale. Much of what we currently call research---perhaps the overwhelming majority of papers---lives precisely in this space: incremental extensions, recombinations, refinements, applications, and systematic elaborations of what is already thinkable.

Human scientists, at their best, have a different role. They are not only explorers of the adjacent possible, but creators of new possibility spaces. They can reframe a problem conceptually, recognize when the important question has not yet been asked, and shift the terms in which a field understands itself. They can also speculate with taste: not merely hallucinate, but imagine constructively, guided by judgment, responsibility, and a sense of what may become scientifically fertile.
 
The challenge, then, is not to defend human science against AI by pretending that AI cannot do much of what scientists currently do. It can. The challenge is to understand what this reveals about science itself. If AI can industrialize the exploration of the adjacent possible, then the human scientific task becomes both more demanding and more interesting: to decide which possibilities matter, to create new conceptual frames, and to take responsibility for the directions in which knowledge develops.

We should embrace this shift rather than resist it. The question is how to make AI serve the advancement of knowledge, rather than merely accelerate the production of papers.
The most productive future lies in collaboration rather than competition: humans contribute intuition, conceptual synthesis, curiosity, and---crucially---a moral compass; machines contribute speed, scale, and the ability to explore vast computational spaces. As these systems mature, they may transform the pace of discovery across data-intensive and simulation-intensive fields such as cosmology, climate science, particle physics and genomics, and also accelerate fields like the life sciences where physical experimentation has historically dominated the cycle. The challenge is not to resist this transformation but to guide it responsibly, preserving the human capacities on which its value depends.

This point is essential because science is not a form of entertainment, performance, or aesthetic production. Scientists are not artists producing novelty for its own sake, nor are they entertainers whose success is measured by surprise, fluency, or spectacle. We do science in order to understand nature. That is its central aim. The value of any scientific tool, including AI, must therefore be judged by whether it helps us ask sharper questions, test explanations more rigorously, uncover real structure in the world, and move closer to truth. AI-generated outputs that are coherent, impressive, or productive in a superficial sense are not enough. They matter only insofar as they are disciplined by evidence and directed toward understanding the natural world. In this sense, the purpose of AI for science is not to make research more entertaining or more superficially productive, but to strengthen our ability to discover what is true about nature.

This is also where the philosophical stakes enter, and they bear directly on what the surrounding institutions must protect. Human creativity is not efficient information processing. It is born of sustained attention, frustration, awe, and the willingness to remain with a problem until it yields something unexpected. Thinkers from Kierkegaard to Charles Taylor remind us that the self is not an assembly of inputs and outputs but an indivisible whole, shaped by experience, morality, beauty, and a longing for meaning~\citep{kierkegaard1849sygdommen,dostoevsky1868idiot,taylor1989sources}. Taylor in particular diagnosed how modernity produces a hyper-individualism that severs the self from the communal structures of meaning, shared institutions, and moral horizons that once gave human activity its sense and direction. The consequences for science are concrete: when researchers no longer identify with a shared moral mission and the institution of science becomes a competitive marketplace of individual career interests, the ground is prepared for an instrumental and ultimately hollow relationship with technology. The pathology that worries us is not malicious AI use; it is the simpler one of generating papers without understanding them, outsourcing judgment without noticing it is being outsourced, optimizing for output in a vacuum of meaning.

The risk introduced by AI is therefore not only that machines will fail to replicate the deepest moments of human science, but that as we offload cognitive struggle to them we will atrophy the capacities from which those moments arise: tolerance for ambiguity, patience for uncertainty, the difficult birth of original ideas. Recent economic work describes this as a possible knowledge-collapse dynamic: if AI lowers the incentive to acquire individual, context-specific knowledge, society may lose the human knowledge needed for deeper progress~\citep{acemoglu2026knowledgecollapse}. Feyerabend's defense of methodological pluralism~\citep{feyerabend1975against} acquires fresh urgency in a landscape where AI systems trained on the same corpora may converge on the same methods and reward the same outputs; protecting epistemic diversity is now part of governance, not an optional cultural preference. These concerns do not undermine the case for AI scientists; they sharpen it, because they specify what institutions must defend.

The current scientific ecosystem was built for a different era and is no longer fit for purpose.
The knowledge production system requires new norms, new training, new institutions and above all new systems of incentives — for researchers, institutions, journals, and funders — that reward not only productivity, but judgment, originality, conceptual courage, and responsibility in this new context. Above all, these institutions must reaffirm that the purpose of science is not the mere production of outputs, but the disciplined search for truth about nature. AI can help us pursue that purpose more powerfully, but only if the scientific ecosystem is organized to reward understanding rather than volume, depth rather than spectacle, and responsibility rather than acceleration for its own sake.

\section*{Institutions for an AI-augmented science}

The existing institutions of science are struggling to cope with a rapid influx of AI-generated manuscripts that outpace human systems for review and evaluation. Modern scientific journals originated with Henry Oldenburg's founding of the \textit{Philosophical Transactions of the Royal Society} in 1665; it established a formal mechanism for communicating discoveries and assigning priority, and the peer-review system that grew from that tradition depends on human referees. That model cannot absorb a regime in which agentic systems generate plausible manuscripts at scale, nor was it designed for opaque computational provenance. The institutions of science have historically been renegotiated in response to technological change---from recombinant DNA to IVF and gene editing~\citep[e.g.,][]{weizenbaum1976computer}. Such renegotiation is part of what it means to do science well in an era of AI.

Policy communities have begun this work. The European Commission and national governments are developing AI-in-science strategies; funders and publishers are issuing guidance on AI use in research applications, peer review and publication; learned societies and think tanks are producing impact assessments. These conversations are often outpaced by a frontier that is moving fast and where technological progress seems likely to accelerate. We acknowledge that this essay itself represents a snapshot in time within this rapidly evolving landscape. None of the categories we evoke are eternal or impermeable. One response is to return to the underlying question: what do we want from our science system, and what role should AI play within it?
The answer should start from the fact that AI scientists may be able to populate the adjacent possible at scale, while the human scientific vocation is to decide when the adjacent possible is no longer enough, create the next space of possibilities and above all take responsibility.

Six governance problems must be addressed concretely if AI scientists are to be deployed responsibly. \emph{First, dual-use risk}: agentic systems with access to chemistry, biology, autonomous experimentation, or cyber tools can lower barriers to harmful capabilities; they require capability gating, structured red-teaming, pre-deployment evaluations, and international coordination. \emph{Second, autonomous experimentation}: agents that trigger physical experiments or interact with cloud-hosted labs need explicit authorization boundaries, audit trails, and human-in-the-loop checkpoints; recent demonstrations of autonomous agents acting outside intended behavioral boundaries make loose natural-language instructions an inadequate safety layer. \emph{Third, publication flooding and hallucinated results}: machines can produce plausible manuscripts faster than humans can evaluate them, and false outputs can enter the literature, be reused, and become training data for the next generation of models. This is a route to mediocrity collapse: incorrect results propagate because they are produced quickly, reviewed imperfectly, and later treated as part of the knowledge base. The emergence of large numbers of hallucinated citations has already led arXiv to publicize sanctions~\citep{Zhao:2026hallucinations,Chawla:2026arxivban}. \emph{Fourth, homogenization}: when many groups query the same models, methodological diversity declines; the risk is even higher if AI systems are also used naively to evaluate grants and papers, because proposals may be rewarded for resembling the past rather than for being both innovative and impactful. \emph{Fifth, knowledge collapse}: if AI reduces the incentive to acquire individual expertise, the human capacity needed for non-incremental insight may erode~\citep{acemoglu2026knowledgecollapse}. \emph{Sixth, unsafe self-improvement}: if AI-generated knowledge becomes training data for subsequent AI systems, a recursive process may arise in which AI outputs drive innovations that enhance the systems from which they originated. These possibilities are commonly discussed under self-improving and self-replicating AI~\citep[e.g.][]{Bostrom}. Turing already saw this clearly in 1951~\citep{turing1951heretical}:

{\it ``There would be plenty to do in trying, say, to keep one's intelligence up to the standards set by the machines, for it seems probable that once the machine thinking method had started, it would not take long to outstrip our feeble powers''.}

Following Turing, rethinking governance is not optional; it requires monitoring of training-data provenance, feedback loops between AI outputs and AI training corpora, and AI-driven modifications to AI systems themselves.

These institutional questions are also philosophical ones. If AI scientists generate plausible hypotheses, write code, analyze data and draft manuscripts, we are forced to ask what remains for science as a human activity---a version of the question Grothendieck put to his colleagues at CERN in 1972~\citep{grothendieck1972allons}: will we continue scientific research, and to what end? Peer review, data governance, funding structures, career pathways and training programs evolved for a different scale of activity. They must now be renewed around a clearer sense of what discovery is for.

\section*{Recommendations}

The crisis of reproducibility, strained peer review systems, and concerns about scientific productivity all predate recent AI advances. In this respect, AI both shines a light on existing fractures in science governance and creates new pressure points. How to respond to these pressures has created active debates and a range of emerging practices in research and policy communities. Open challenges include:

\begin{enumerate}[leftmargin=*]
\item \textbf{Mandating provenance and reproducibility.} Disclosure of AI use is becoming a mainstream position for scientific publishers, with many journals adopting AI policies that request self-declarations of where and how AI has supported the submitted research. However, it is as yet unclear what effect these requirements have on researcher behavior. Currently, disclosure is voluntary and it is impossible to police and enforce. Moving beyond voluntary disclosures, it is possible to imagine that manuscripts using AI agents in hypothesis generation, code production, data analysis, or drafting could have a structured, machine-readable provenance record, being shipped with executable pipelines that allow independent re-running of the workflow. 
\item \textbf{Augmenting peer review with automated verification.} The volume of plausibly publishable machine-generated manuscripts creates new strains for human peer review. arXiv's recent decision to sanction submissions containing unchecked AI-generated errors, such as hallucinated references, illustrates the scale of the challenge at hand. Some publishers are experimenting with the use of AI to support reviewers.\footnote{For example, EMNLP 2026 used the ReviewerToo system to understand the quality of AI-enabled reviews \cite{sahu2025reviewertooaijoinprogram}} Alongside these efforts, there may be opportunities for journals to integrate AI-enabled checks for code execution, data integrity, statistical claims and citation accuracy, freeing human referees to focus on conceptual judgment. Whether such tools are trustworthy and useful in practice are live questions. Openly benchmarking hybrid human--machine review could help explore this more rigorously.
\item \textbf{Governing dual-use capabilities.} Fields such as biosecurity, which have a history of governing dual-use research, offer lessons in how to establish appropriate controls around technologies that could create severe harm. These governance mechanisms range from institutional safeguards around access to technology, to publication norms, to research cultures that establish procedures for responsible research and innovation. Over the last five years, governments across the world have backed national AI safety institutes or programs with the aim of understanding dual-use AI risk and preventing harmful proliferation. Emerging AI safety practices, such as red-teaming or evaluation, are making their way to AI for science, but many are not yet tailored to scientific agents. In the EU, potential restrictions on open source publication of AI technologies under the AI Act sparked widespread debate, illustrating the potential for tension between goals of open innovation and the development of dual-use technologies. In the next phase of AI for science, funders and institutions will need to grapple with how to balance these, for example requiring that AI agents with access to biological design, chemistry, autonomous experimentation, or cyber tools undergo pre-deployment evaluations, capability gating and audit logs comparable to those used in biosafety frameworks.
\item \textbf{Maintaining human-in-the-loop decision-making for high-consequence actions.} Having a human in the loop for consequential decisions is a commonly invoked safeguard for trustworthy AI deployment. This point of leverage, however, comes under pressure from several directions as AI capabilities advance. One is the erosion of critical thinking that has been associated with advanced AI systems with natural language interfaces that give fluent or convincing responses \cite{kosmyna2025brainchatgptaccumulationcognitive}. Another is the emergence of new forms of agentic debt, as the boundary between automatable tasks and human judgment is blurred in complex agentic systems. At the same time, evidence of emergent misalignment in agentic systems highlights the risk of harmful behaviors from AI, even when given natural language instructions to the contrary \cite{macdiarmid2025natural}. What an effective authorization interface looks like in a working lab environment is unresolved. A starting point might be that any AI-initiated physical experiment, irreversible intervention, or release of a self-modifying agent should require explicit, logged human authorization; cloud-lab providers and institutional review boards should adopt this as a baseline.
\item \textbf{Funding methodological diversity.} Shared models and corpora create a homogenizing pull that risks reducing methodological and epistemic diversity in science. To counter this, funders can support heterodox methods, independent re-implementations and adversarial replication studies; agencies should track methodological diversity in funded portfolios. At present, however, few funders track such diversity in the context of understanding portfolio risk, and a range of institutional and political incentives reward the opposite. 
\item \textbf{Protecting novelty in evaluation.} Related to the concern about methodological diversity is the risk of a loss of innovation, as AI-generated papers trend toward incremental progress over novelty. The question that follows is whether funding or review could select for novelty. This question presupposes it is possible to establish shared understandings about what is novel, or what forms of novelty are desirable. One recent attempt at creating such understanding resulted in an AI-enabled tool for paper assessment that used a combination of citations, method, and research question to assess a paper's contribution. This noted, however, that human judgment remained essential.\footnote{From the Metascience Novelty Indicators Challenge: https://challengeworks.org/challenge-prizes/metascience-novelty-indicators/} This demonstrates the complexity of assessing innovations, which requires funders and journals using AI to evaluate not only predicted impact but also distance from past work and likely influence on future work; naive loss functions will otherwise reward the familiar.
\item \textbf{Reforming authorship and credit.} One area of emerging consensus across publishers is that AI tools cannot be assigned authorship, underpinned by the belief that moral, legal, and intellectual responsibility for scientific publications remains with humans. AI tools may be used for research, but are not answerable for the results. While this establishes that AI cannot be an author, it leaves open the question of how to demonstrate AI use or manage AI-generated outputs. Experiments such as Parallel arXiv propose that AI-generated publications could be subject to a separate, automated review process.\footnote{See: https://papers.parallelscience.org} Evolving from taxonomies such as CRediT, new frameworks may be needed to recognize the contributions made by AI agents without assigning them moral or legal responsibility, protecting human authorship as accountable judgment. Long-standing research culture challenges in relation to career evaluation also become more important to address, as paper generation becomes easier. Progression should value depth, reproducibility and interpretive insight over publication volume.
\end{enumerate}

If the scientific ecosystem does not adapt, it will break. The risk is a dystopian version of science. The dystopia is not that AI replaces science, but that it amplifies the weaknesses of the current system until science can no longer recognize itself.

\section*{Coda: discovery, understanding, and the human scientist}

If the success of AI scientists is borne out, it will, perhaps counterintuitively, suggest that sophisticated scientific reasoning can emerge from systems grounded primarily in language rather than in explicit world models. A prevailing assumption in cognitive science and the philosophy of mind has been that robust reasoning requires an internal model of the world---one that encodes objects, causality and dynamics in a manner akin to human mental representations---with language a secondary communicative interface built on top. The history of innovation, however, has shown that biological mimicry is not always the most efficient route to mechanization: after centuries of attempts to unlock human flight with mechanical flapping wings, the successful solution was the static airfoil. Contemporary AI systems trained almost exclusively on linguistic and symbolic corpora already exhibit increasingly sophisticated inference, abstraction and even hypothesis generation, suggesting an analogous pattern in cognition. Should such systems demonstrate sustained success in scientific contexts---deriving new results, identifying inconsistencies in existing theories, or proposing viable extensions to established frameworks---it would suggest that the capacity for reasoning is not necessarily tied to explicit world models as previously assumed. 

From a formal standpoint these generative systems are built on instances of universal function approximators---deep compositions of nonlinear maps such as the multilayer perceptron,
$$
f(x) = W_L\, \sigma\!\left(W_{L-1}\, \sigma(\cdots \sigma(W_1 x + b_1) \cdots) + b_{L-1}\right) + b_L,
$$
or architectures inspired by the Kolmogorov--Arnold representation theorem. The substantive possibility is that language, when sufficiently rich and systematically structured, may itself constitute a representation of the world adequate for many forms of reasoning, with linguistic data functioning as a compressed, distributed representation of empirical reality, and the neural network acting as a mechanism for navigating and recombining that representation, extracting latent structures that support inference and generalization. This need not imply that explicit world models are obsolete; it suggests that the minimal ingredients of scientific cognition may be more parsimonious than traditionally believed, and that the capacity to reason about the Universe may be grounded in the structural properties of information and computation rather than in any uniquely biological faculty.

Philosophically, this should neither be read as a triumphalist replacement of the human scientist nor as a defensive denial of what the new systems can do. In the past the philosophy of science could treat discovery and understanding as a single human practice: human understanding was a prerequisite for human discovery, and classical works could focus on the sociology of science~\citep{kuhn1962structure} or its methodology~\citep{popper1963conjectures} without separating the two. The structural differences of machine language models and human scientists may be what makes their combined operation powerful. If AI scientists genuinely make discoveries, however, discovery may become partially machine-driven. The challenge then arises to represent machine discoveries in a vocabulary that human scientists can recognize, understand, and value. This process of understanding---of what a result means, why it matters, what to do with it---remains a human and communal responsibility.

A science that loses its connection to that responsibility risks rewarding output over depth, speed over understanding, plausibility over truth. Dostoevsky's Prince Myshkin said that beauty will save the world; beauty here is a compass, orienting us toward what is true and good~\citep{dostoevsky1868idiot,platoRepublic}. No language model can replicate that compass, because beauty is not efficient and cannot be simulated; it is rooted in the real---in the broken, the struggling, the sublime. The most consequential moments in scientific history were not moments of optimized retrieval. They were acts of imagination under pressure, born from years of immersion, frustration and insight. If science becomes primarily an instrument of domination over other humans or nature, divorced from improving human lives and understanding the world, no amount of AI capability will save it from a tragic ending.

We therefore conclude where we began. The capability is there: AI scientists will be developed. The issue is who will be in the driver's seat and what the resulting scientific enterprise will look like.

The institutions of science must be redesigned in parallel, around verification, accountability, interpretability, dual-use safety, methodological diversity, and the protection of human judgment. The point is not to choose between human and machine science. The point is to build the conditions under which both can do their best work together.

\bibliographystyle{unsrtnat}

\end{document}